\documentclass[runningheads]{llncs}

\usepackage{graphicx}
\usepackage{amsmath,amssymb}
\usepackage{mathtools}
\usepackage{xspace}
\usepackage{booktabs}
\usepackage{xcolor}
\usepackage{colortbl}
\usepackage{multirow}
\usepackage{array}
\usepackage{hyperref}
\usepackage{url}
\usepackage{changepage}

\usepackage{booktabs}
\usepackage{longtable}
\usepackage{nicefrac}
\usepackage{amssymb}
\usepackage{amsmath}
\usepackage{xspace}
\usepackage{mathtools}
\usepackage{diagbox}
\usepackage{graphics}
\usepackage{rotating}
\usepackage[algo2e,ruled,vlined,linesnumbered]{algorithm2e}
\newcommand{\assign}{\leftarrow}
\usepackage{graphicx}
\usepackage{subfigure}
\usepackage{ifthen}
\usepackage{wrapfig}
\usepackage{balance} 
\usepackage{epsfig}
\usepackage{xcolor, colortbl}
\usepackage[normalem]{ulem}

\usepackage{array}
\newcolumntype{L}[1]{>{\raggedright\let\newline\\\arraybackslash\hspace{0pt}}m{#1}}
\newcolumntype{C}[1]{>{\centering\let\newline\\\arraybackslash\hspace{0pt}}m{#1}}
\newcolumntype{R}[1]{>{\raggedleft\let\newline\\\arraybackslash\hspace{0pt}}m{#1}}

\renewcommand{\epsilon}{\varepsilon}

\newcommand{\mlga}{$(\mu+\lambda)$~GA\xspace}

\newcommand{\om}{\textsc{OneMax}\xspace}
\newcommand{\onemax}{\om}
\newcommand{\lo}{\textsc{LeadingOnes}\xspace}
\newcommand{\leadingones}{\lo}

\begin{document}
\title{Benchmarking a $(\mu+\lambda)$ Genetic Algorithm with Configurable Crossover Probability}
\titlerunning{Benchmarking a $(\mu+\lambda)$ GA with Configurable Crossover Probability}
\author{Furong Ye\inst{1} \and Hao Wang\inst{2} \and Carola Doerr\inst{2} \and Thomas B\"ack\inst{1}}

\institute{
LIACS, Leiden University, Leiden, The Netherlands 
\email{\{f.ye,t.h.w.baeck\}@liacs.leidenuniv.nl} 
\and
Sorbonne Universit\'e, CNRS, LIP6, Paris, France
\email{\{hao.wang,carola.doerr\}@lip6.fr} 
}

\maketitle

\begin{abstract}
We investigate a family of $(\mu+\lambda)$ Genetic Algorithms (GAs) which creates offspring either from mutation or by recombining two randomly chosen parents. By scaling the crossover probability, we can thus interpolate from a fully mutation-only algorithm towards a fully crossover-based GA. We analyze, by empirical means, how the performance depends on the interplay of population size and the crossover probability.

Our comparison on 25 pseudo-Boolean optimization problems reveals an advantage of crossover-based configurations on several easy optimization tasks, whereas the picture for more complex optimization problems is rather mixed. Moreover, we observe that the ``fast'' mutation scheme with its are power-law distributed mutation strengths outperforms standard bit mutation on complex optimization tasks when it is combined with crossover, but performs worse in the absence of crossover. 

We then take a closer look at the surprisingly good performance of the crossover-based $(\mu+\lambda)$ GAs on the well-known LeadingOnes benchmark problem. We observe that the optimal crossover probability increases with increasing population size $\mu$. At the same time, it decreases with increasing problem dimension, indicating that the advantages of the crossover are not visible in the asymptotic view classically applied in runtime analysis. We therefore argue that a mathematical investigation for fixed dimensions might help us observe effects which are not visible when focusing exclusively on asymptotic performance bounds.

\end{abstract}

\sloppy{
\section{Introduction}

Classic evolutionary computation methods build on two main variation operators: \textit{mutation} and \textit{crossover}. While the former can be mathematically defined as unary operators (i.e., families of probability distributions that depend on a single argument), crossover operators sample from distributions of higher arity, with the goal to ``recombine'' information from two or more arguments. 

There is a long debate in evolutionary computation about the (dis-)advantages of these operators, and about how they interplay with each other~\cite{MitchellHF93,spears1993crossover}. In lack of generally accepted recommendations, the use of these operators still remains a rather subjective decision, which in practice is mostly driven by users' experience. Little guidance is available on which operator(s) to use for which situation, and 
how to most efficiently interleave them. The question how crossover can be useful can therefore be seen as far from being solved. 

Of course, significant research efforts are spent to shed light on this question, which is one of the most fundamental ones that evolutionary computation has to offer. 
While in the early years of evolutionary computation (see, for example, the classic works~\cite{DeJong75,Goldberg89,baeck96}) crossover seems to have been widely accepted as an integral part of an evolutionary algorithm, we observe today two diverging trends. 
Local search algorithms such as GSAT~\cite{GSAT} for solving Boolean satisfiability problems, or such as the general-purpose Simulated Annealing~\cite{SA83} heuristic, are clearly very popular optimization methods in practice -- both in academic and in industrial applications. These purely mutation-based heuristics are nowadays more commonly studied under the term \emph{stochastic local search}, which forms a very active area of research. 
Opposed to this is a trend to reduce the use of mutation operators, and to fully base the iterative optimization procedure on recombination operators; see~\cite{Whitley19crossover} and references therein. 
However, despite the different recommendations, these opposing positions find their roots in the same problem: we hardly know how to successfully dovetail mutation and crossover. 

In addition to large bodies of empirical works aiming to identify useful combinations of crossover and mutation~\cite{DeJong75,Murata96,Yoon22,ELSAYED20111877},  

the question how (or whether) crossover can be beneficial has also always been one of the most prominent problems in \textit{runtime analysis}, the research stream aiming at studying evolutionary algorithms by mathematical means
~\cite{DBLP:conf/gecco/WatsonJ07,kotzing2011crossover,Jansen2002Crossover,Lehre2010UIOCrossover,jansen2005real,DoerrJKNT13tcs,Sudholt05,doerr2012crossover,DoerrDE15,sudholt2017crossover,CorusO18,CorusO19,PintoD18,WhitleyVHM18,DBLP:conf/gecco/NeumannORS11,dang2017escaping}
, most of these results focus on very particular algorithms or problems, and are not (or at least not easily) generalizable to more complex optimization tasks. 

\paragraph{Our Results}

In this work, we study a simple variant of the \mlga which allows us to conveniently scale the relevance of crossover and mutation, respectively, via a single parameter. More precisely, our algorithm is parameterized by a crossover probability $p_c$, which is the probability that we generate in the reproduction step an offspring by means of crossover. The offspring is generated by mutation otherwise, so that $p_c=0$ corresponds to the mutation-only $(\mu+\lambda)$~EA, whereas for $p_c=1$ the algorithm is entirely based on crossover. Note here that we \textit{either} use crossover \textit{or} mutation, so as to better separate the influence of the two operators.

We first study the performance of different configurations of the \mlga on 25 pseudo-Boolean problems (the 23 functions suggested in~\cite{doerr2019benchmarking}, a concatenated trap problem, and random NK landscape instances). We observe that the algorithms using crossover perform significantly better on some simple functions as \onemax (F1) and \leadingones (F2), but also on some problems that are considered hard, e.g., the 1-D Ising model (F19). 

We then look more closely into the performance of the algorithm on a benchmark problem intensively studied in runtime analysis: \leadingones, the problem of maximizing the function $f:\{0,1\}^n \rightarrow [0..n], x \mapsto \max\{i \in [0..n] \mid \forall j \le i: x_j=1\}$. We observe some very interesting effects, that we believe may motivate the theory community to look at the question of usefulness of crossover from a different angle. More precisely, we find that, against our intuition that uniform crossover cannot be beneficial on \leadingones, the performance of the \mlga on \leadingones improves when $p_c$ takes values greater than $0$ (and smaller than 1), see Fig.~\ref{fig:runtime-mumu-LO-sbm}. The performances are quite consistent, and we can observe clear patterns, such as a tendency for the optimal value of $p_c$ (displayed in Tab.~\ref{tab:pc-opt-LO-sbm}) to increase with increasing $\mu$, and to decrease with increasing problem dimension. The latter effect may explain why it is so difficult to observe benefits of crossover in theoretical work: they disappear with the asymptotic view that is generally adopted in runtime analysis. 

We have also performed similar experiments on \onemax (see our project data~\cite{FURONG2020_3752064}), but the good performance of the \mlga configurations using crossover is less surprising for this problem, since this benefit has previously been observed for genetic algorithms that are very similar to the \mlga; see~\cite{sudholt2017crossover,PintoD18,CorusO18,CorusO19} for examples and further references. In contrast to a large body of literature on the benefit of crossover for solving \onemax, we are not aware of the existence of such results for \leadingones, apart from the highly problem-specific algorithms developed and analyzed in~\cite{AfshaniADLMW12,DoerrW11EA}. 

We hope to promote with this work 
(1) runtime analysis for fixed dimensions, 
(2) an investigation of the advantages of crossover on \leadingones, and 
(3) the \mlga as a simplified model to study the interplay between problem dimension, population sizes, crossover probability, and mutation rates.

\section{Algorithms and Benchmarks}
\label{sec:setup}
We describe in this section our \mlga framework (Sec.~\ref{sec:alogs}) and the benchmark problems (Sec.~\ref{sec:problems}). Since in this paper we can only provide a glimpse on our rich data sets, we also summarize in Sec.~\ref{sec:data} which data the interested reader can find in our repository~\cite{FURONG2020_3752064}. 

\subsection{A Family of $(\mu+\lambda)$~Genetic Algorithms}
\label{sec:alogs}
Our main objective is to study the usefulness of crossover for different kinds of problems. To this end, we investigate a meta-model, which allows us to easily transition from a mutation-only to a crossover-only algorithm. Alg.~\ref{alg:GA} presents this framework, which, for ease of notation, we refer to as the family of the \mlga in the following. 

The \mlga initializes its population uniformly at random (u.a.r., lines 1-2). In each iteration, it creates $\lambda$ offspring (lines 6--16). For each offspring, we first decide whether to apply crossover (with probability $p_c$, lines 8--11) or whether to apply mutation (otherwise, lines 12--15). Offspring that differ from their parents are evaluated, whereas offspring identical to one of their parents inherit this fitness value without function evaluation (see~\cite{PintoD18PPSN} for a discussion). The best $\mu$ of parent and offspring individuals form the new parent population of the next generation (line~17). 

Note the unconventional use of \emph{either} crossover \emph{or} mutation. As mentioned, we consider this variant to allow for a better attribution of the effects to each of the operators. Moreover, note that in Alg.~\ref{alg:GA} we decide for each offspring individually which operator to apply. We call this scheme the \textbf{\mlga with offspring-based variator choice}. We also study the performance of the \textbf{\mlga with population-based variator choice}, which is the algorithm that we obtain from Alg.~\ref{alg:GA} by swapping lines~7 and~6. 

\begin{algorithm2e}[t]
\textbf{Input:} Population sizes $\mu$, $\lambda$, crossover probability $p_c$, mutation rate $p$\;
\textbf{Initialization:} 
    \lFor{$i=1,\ldots,\mu$}{sample $x^{(i)} \in \{0,1\}^n$ uniformly at random (u.a.r.), and evaluate $f(x^{(i)})$}
		Set $P = \{x^{(1)},x^{(2)},..., x^{(\mu)}\}$ \;
	\textbf{Optimization:}
	\For{$t=1,2,3,\ldots$}{
	    $P' \leftarrow \emptyset$\;
	    \For{$i=1,\ldots,\lambda$}{
	        Sample $r \in [0,1]$ u.a.r. \;
	        \eIf{$r \le p_c$}{
	            select two individuals $x,y$ from $P$ u.a.r. (w/ replacement)\;
	            $z^{(i)} \assign \text{Crossover}(x,y)$\;
	            \lIf{$z^{(i)} \notin \{x,y\}$}{evaluate $f(z^{(i)})$ \textbf{else} infer $f(z^{(i)})$ from parent}
	        }{
	            select an individual $x$ from $P$ u.a.r.\;
	            $z^{(i)} \assign \text{Mutation}(x)$\;
	            \lIf{$z^{(i)} \neq x$}{evaluate $f(z^{(i)})$ \textbf{else} infer $f(z^{(i)})$ from parent}
	        }
	        $P'\leftarrow P'\cup\{z^{(i)}\}$\;
	    }
        $P$ is updated by the best $\mu$ points in $P \cup P'$ (ties broken u.a.r.)\;
	}
\caption{A Family of $(\mu+\lambda)$~Genetic Algorithms}
\label{alg:GA}
\end{algorithm2e}

We study three different crossover operators, \textit{one-point crossover}, \textit{two-point crossover}, and \textit{uniform crossover}, and two different mutation operators, \emph{standard bit mutation} and the \emph{fast mutation} scheme suggested in~\cite{doerr2017fast}. These variation operators are briefly described as follows.\\
- \emph{One-point crossover}: a crossover point is chosen from $[1..n]$ u.a.r. and an offspring is created by copying the bits from one parent until the crossover point and then copying from the other parent for the remaining positions.\\
-  \emph{Two-point crossover}: similarly, two different crossover points are chosen u.a.r. and the copy process alternates between two parents at each crossover point. \\
-  \emph{Uniform crossover} creates an offspring by copying for each position from the first or from the second parent, chosen independently and u.a.r.  \\
- \emph{Standard bit mutation:} a mutation strength $\ell$ is sampled from the conditional binomial distribution Bin$_{>_0}(n,p_m)$, which assigns to each $k$ a probability of $\binom{n}{k}p^k(1-p)^{n-k}/(1-(1-p)^n)$~\cite{PintoD18PPSN}. Thereafter, $\ell$ distinct positions are chosen u.a.r. and the offspring is created by first copying the parent and then flipping the bits in these $\ell$ positions. In this work, we restrict our experiments to the standard mutation rate $p=1/n$. Note, though, that this choice is not necessarily optimal, as in particular the results in~\cite{BottcherDN10,sudholt2017crossover} and follow-up works demonstrate.\\
- \emph{Fast mutation}~\cite{doerr2017fast}: operates similarly to standard bit mutation except that the mutation strength $\ell$ is drawn from a power-law distribution: $\operatorname{Pr}[L=\ell]=(C_{n/2}^{\beta})^{-1}\ell^{-\beta}$ with $\beta = 1.5$ and $C_{n/2}^{\beta}=\sum_{i=1}^{n/2} i^{-\beta}$.

\subsection{The IOHprofiler Problem Set}
\label{sec:problems}

To test different configurations of the \mlga, we first perform an extensive benchmarking on the problems suggested in~\cite{doerr2019benchmarking}, which are available in the IOHprofiler benchmarking environment~\cite{IOHprofiler}. This set contains $23$ real-valued pseudo-Boolean test problems: \textbf{F1 and F4-F10:} \onemax (F1) and W-model extensions (F4-10), \textbf{F2 and F11-F17:} \leadingones (F2) and W-model extensions (F11-17), \textbf{F3:} Linear function $f(\mathbf{x})=\sum_{i=1}^n i{x_i}$, \textbf{F18:} Low Autocorrelation Binary Sequences (LABS), \textbf{F19-21:} Ising Models, \textbf{F22:} Maximum Independent Vertex Set (MIVS), and \textbf{F23:} N-Queens (NQP).

We recall that the W-model, originally suggested in~\cite{weise2018difficult} and extended in~\cite{doerr2019benchmarking}, is a collection of perturbations that can be applied to a base problem in order to calibrate its features, such as its neutrality, its epistasis, and its ruggedness. We add to the list of~\cite{doerr2019benchmarking} the following two problems:\\
 \textbf{F24:} Concatenated Trap (CT) is defined by partitioning a bit-string into segments of length $k$ and concatenating $m=n/k$ trap functions that takes each segment as input. The trap function is defined as follows: $f_k^{\text{trap}}(u) = 1$ if the number $u$ of ones satisfies $u = k$ and $f_k^{\text{trap}}(u) = \frac{k-1-u}{k}$ otherwise. We use $k=5$ in our experiments.\\
 \textbf{F25:} Random NK landscapes (NKL). The function values are defined as the average of $n$ sub-functions $F_i \colon [0..2^{k+1}-1] \rightarrow \mathbb{R}, i \in [1..n]$, where each component $F_i$ only takes as input a set of $k \in [0..n-1]$ bits that are specified by a neighborhood matrix. In this paper, $k$ is set to $1$ and entries of the neighbourhood matrix are drawn u.a.r.~in $[1..n]$. The function values of $F_i$'s are sampled independently from a uniform distribution on $(0, 1)$.

Note that the IOHprofiler problem set provides for each problem several problem instances, which all have isomorphic fitness landscapes, but different problem representations. In our experiments we only use the first instance of each problem (seed $1$). For the mutation-based algorithms and the ones using uniform crossover, the obtained results generalize to all other problem instances. For algorithms involving one- or two-point crossover, however, this is not the case, as these algorithms are not unbiased (in the sense of Lehre and Witt~\cite{LehreW12}). 

\subsection{Data Availability} 
\label{sec:data}

Detailed results for the different configurations of the \mlga are available in our data repository at~\cite{FURONG2020_3752064}. In particular, we host there data for the IOHprofiler experiments
(36 algorithms, 25 functions, 5 dimensions $\le 250$, 100 independent runs) and for the $(\mu+\lambda)$~GA on \onemax and on \leadingones for all of the following $5544$ parameter combinations: 
     $n \in \{64,100,150,200,250,500\}$ (6 values), 
     $\mu \in \{2,3,5,8,10,20,30,...,100\}$ (14 values), 
     $\lambda \in \{1, \lceil \mu/2 \rceil, \mu\}$ (3 values),
     $p_c \in \{0.1 k \mid k \in [0..9]\}\cup\{0.95\}$ (11 values), two mutation operators (standard bit mutation and fast mutation). In these experiments on \onemax and \leadingones, the crossover operator is fixed to uniform crossover.  

A detailed analysis of these results, for example using IOHprofiler or using HiPlot~\cite{HiPlot} may give additional insights into the dependence of the overall performance on the parameter setting

\begin{figure*}[t]
 \includegraphics[width=\linewidth, trim = 0mm 5mm 0mm 0mm, clip]{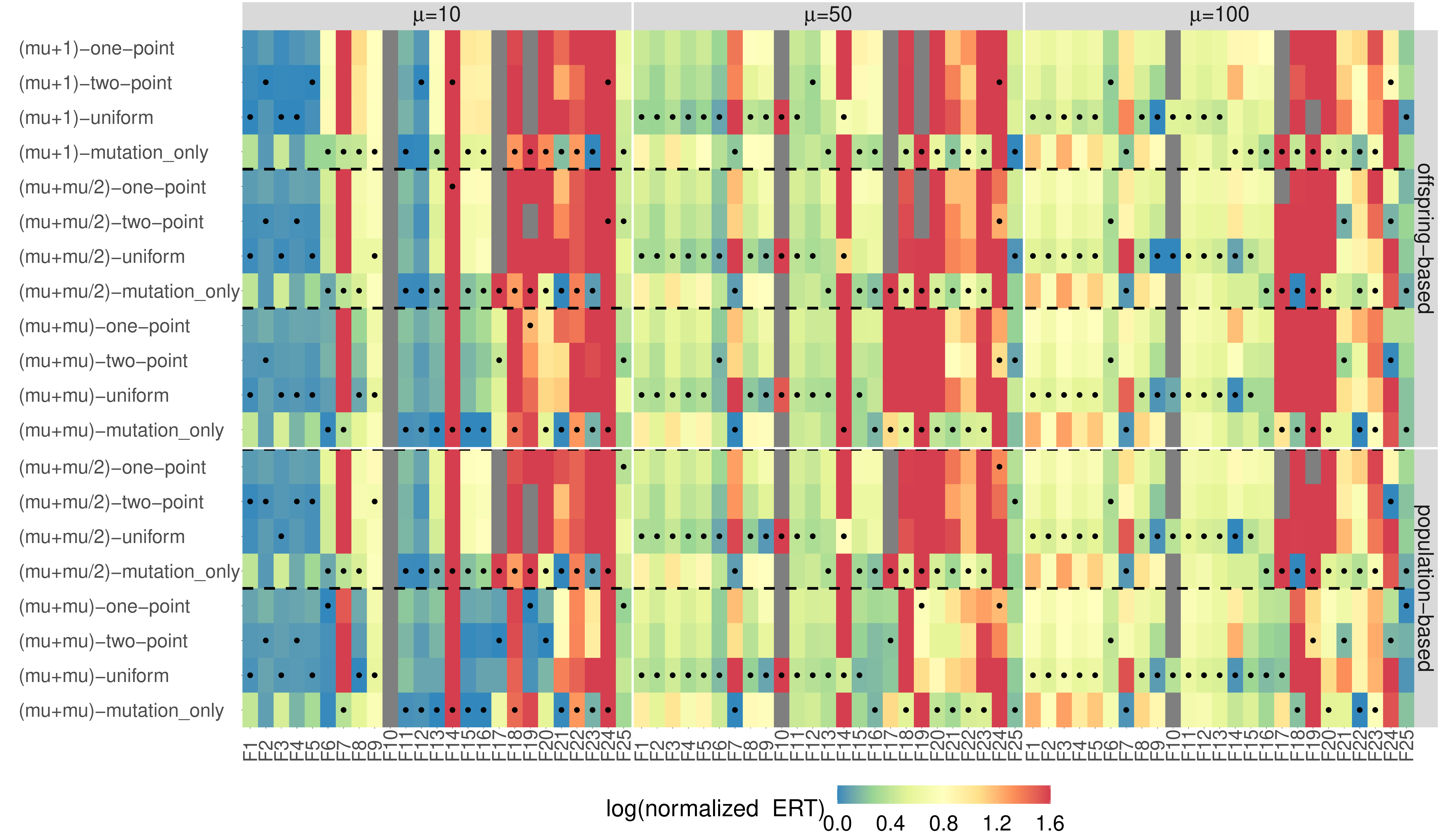}
\caption{Heat map of normalized ERT values of the \mlga with offspring-based (top part) and population-based (bottom part) variator choice for the $100$-dimensional benchmark problems, computed based on the target values specified in Table~\ref{tab:targets}. The crossover probability $p_c$ is set to $0.5$ for all algorithms except the mutation-only ones (which use $p_c = 0$). The  displayed values are the the quotient of the ERT and $\text{ERT}_\text{best}$, the ERT achieved by the best of all displayed algorithms. These quotients are capped at $40$ to increase interpretability of the color gradient in the most interesting region. The three algorithm groups -- the $(\mu+1)$, the $(\mu+\lceil\mu/2\rceil)$, and the $(\mu+\mu)$~GAs -- are separated by dashed lines. A dot indicates the best algorithm of each group of four. A grey tile indicates that the \mlga configuration failed, in all runs, to find the target value within the given budget.
}
 \label{fig:ERT-all}
\end{figure*}

\begin{figure*}[t]
 \includegraphics[width=\linewidth, trim = 0mm 5mm 0mm 0mm, clip]{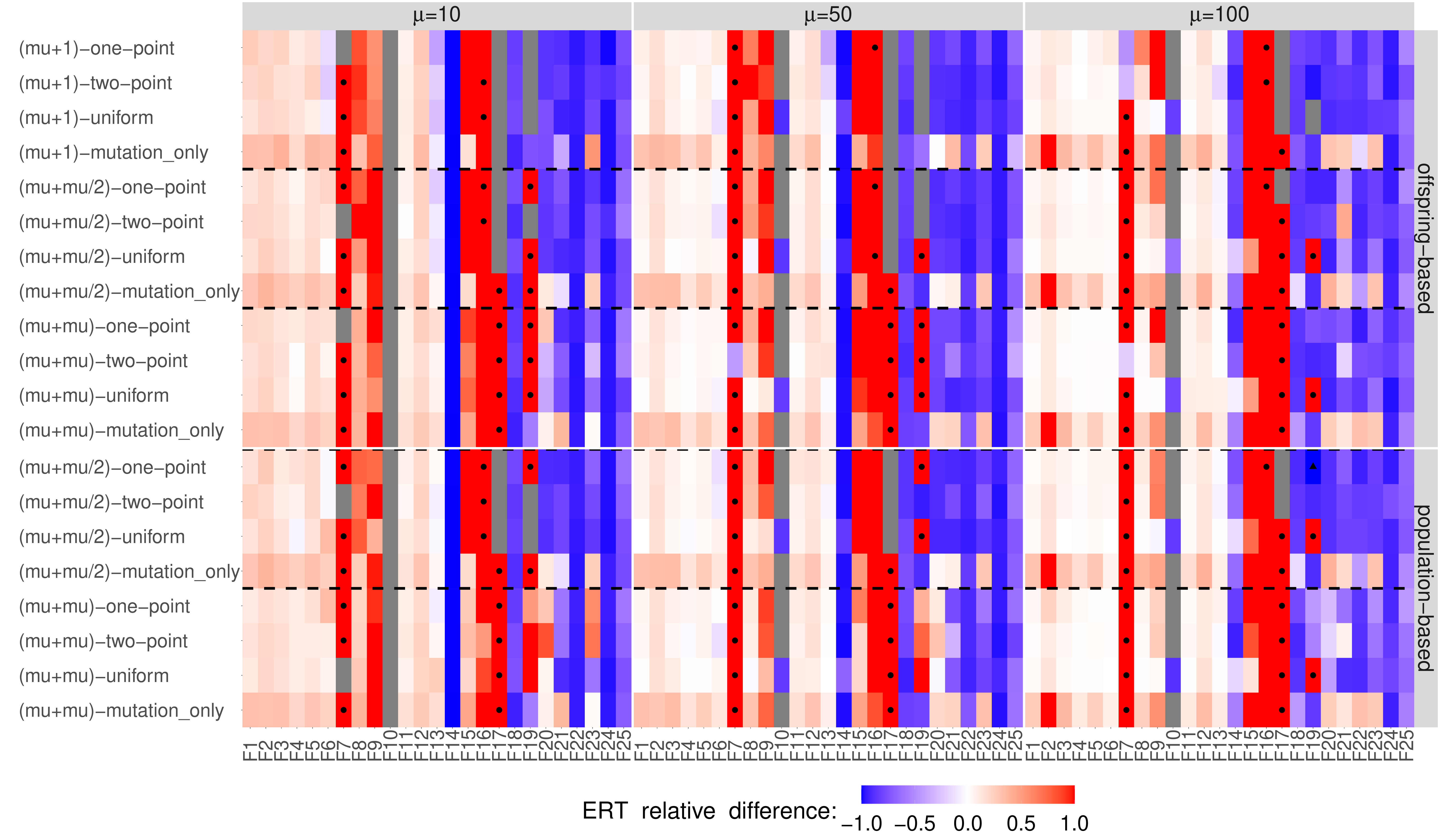}
\caption{Heat map comparing standard bit mutation (\text{sbm}) with fast mutation on the $25$ problems from Sec.~\ref{sec:problems} in dimensions $n=100$. Plotted values are $(\text{ERT}_{\text{fast}}- \text{ERT}_{\text{sbm}})/{\text{ERT}_{\text{sbm}}}$, for ERTs computed wrt the target values specified in Table~\ref{tab:targets}. $p_c$ is set to $0.5$ for all crossover-based algorithms. Values are bounded in $[-1,1]$ to increase visibility of the color gradient in the most interesting region. A black dot indicates that the \mlga with fast mutation failed in all runs to find the target with the given budget; the black triangle signals failure of standard bit mutation, and a gray tile is chosen for settings in which the \mlga failed for both mutation operators.
}
 \label{fig:ERT-sbm-vs-fast}
\end{figure*}

\section{Results for the IOHprofiler Problems}
\label{sec:ioh}

\begin{table}[t]
\small
\setlength{\tabcolsep}{2.5pt}
\centering
\begin{tabular}{r|r|r|r|r|r|r|r|r|r|r|r|r|r}
\hline
$F1$ & $F2$ & $F3$ & $F4$ & $F5$ & $F6$ & $F7$ & $F8$ & $F9$ & $F10$ & $F11$ & $F12$ & $F13$ & $F14$   \\\hline
100 & 100 & 5050 & 50 & 90 & 33 & 100 & 51 & 100 & 100 & 50 & 90 & 33 & 7 \\ \hline
\end{tabular}
\begin{tabular}{r|r|r|r|r|r|r|r|r|r|r}
\hline
$F15$ & $F16$ & $F17$ & $F18$ & $F19$ & $F20$ & $F21$ & $F22$ & $F23$ & $F24$ & $F25$ \\\hline
51 & 100 & 100 & 4.215852 & 98 & 180 & 260 & 42 & 9 & 17.196 & -0.2965711 \\ \hline
\end{tabular}
\caption{\small Target values used for computing the ERT value in Fig.~\ref{fig:ERT-all}.} 
\vspace{-5ex}
\label{tab:targets}
\end{table}

In order to probe into the empirical performance of the \mlga, we test it on the 25 problems mentioned in Sec.~\ref{sec:problems}, with a total budget of $100n^2$ function evaluations. We perform $100$ independent runs of each algorithm on each problem. A variety of parameter settings are investigated: (1) all three crossover operators described in Sec.~\ref{sec:setup} (we use $p_c =0.5$ for all crossover-based configurations), (2) both mutation variator choices, (3) $\mu \in \{10, 50, 100\}$, and (4) $\lambda \in \{1, \lceil\mu/2\rceil, \mu\}$. 

In Fig.~\ref{fig:ERT-all}, we highlight a few basic results of this experimentation for $n=100$, where the mutation operator is fixed to the standard bit mutation. More precisely, we plot in this figure the normalized expected running time (ERT), where the normalization is with respect to the best ERT achieved by any of the algorithms for the same problem. 
Table~\ref{tab:targets} provides the target values for which we computed the ERT values. For each problem and each algorithm, we first calculated the 2\% percentile of the best function values. We then selected the largest of these percentiles (over all algorithms) as target value.

On the \onemax-based problems F1, F4, and F5, the \mlga outperforms the mutation-only GA, regardless of the variator choice scheme, the crossover operator, and the setting of $\lambda$. 
When looking at problem F6, we find out that when $\mu=10$ the mutation-only GA surpasses most of \mlga variants except the population-based $(\mu+\mu)$ GA with one-point crossover.
On F8-10, the \mlga takes the lead in general, whereas it cannot rival the mutation-only GA on F7. Also, only the configuration with uniform crossover can hit the optimum of F10 within the given budget. 

On the linear function F3 we observe a similar behavior as on \onemax. On \leadingones (F2), the \mlga outperforms the mutation-only GA again for $\mu\in\{50,100\}$ while for $\mu = 10$ the mutation-only GA becomes superior with one-point and uniform crossovers. On F11-13 and F15-16 (the W-model extensions of \leadingones), the mutation-only GA shows a better performance than the \mlga with one-point and uniform crossovers and this advantage becomes more significant when $\mu=10$. On problem F14, that is created from \leadingones using the same transformation as in F7, the mutation-only GA is inferior to the \mlga with uniform crossover. 

On problems F18 and F23, the mutation-only GA outperforms the \mlga for most parameter settings. On F21, the \mlga with two-point crossover yields a better result when the population size is larger (i.e., $\mu=100$) while the mutation-only GA takes the lead for $\mu=10$. On problems F19 and F20, the $(\mu+\mu)$~GA with the population-based variator choice significantly outperforms all other algorithms, whereas it is substantially worse for the other parameter settings. On problem F24, the $(\mu+\mu/2)$~GA with two-point crossover achieves the best ERT value when $\mu=100$. None of the tested algorithms manages to solve F24 with the given budget. The target value used in Fig.~\ref{fig:ERT-all} is $17.196$, which is below the optimum $20$.  On problem F25, the mutation-only GA and the \mlga are fairly comparable when $\mu\in\{10, 50\}$. Also, we observe that the population-based $(\mu+\mu)$~GA outperforms the mutation-only GA when $\mu=100$.

In general, we have made the following observations: (1) on problems F1-6, F8-9, and F11-13, all algorithms obtain better ERT values with $\mu=10$. On problems F7, F14, and F21-25, the \mlga benefits from larger population sizes, i.e., $\mu=100$; (2) The $(\mu+\mu)$~GA with uniform crossover and the mutation-only GA outperform the $(\mu+\lceil\mu/2\rceil)$~GA across all three settings of $\mu$ on most of the problems, except F10, F14, F18, and F22. For the population-based variator choice scheme, increasing $\lambda$ from one to $\mu$ improves the performance remarkably on problems F17-24. Such an improvement becomes negligible for the offspring-based scheme; (3) Among all three crossover operators, the uniform crossover often surpasses the other two on \onemax, \leadingones, and the W-model extensions thereof.

To investigate the impact of mutation operators on GA, we plot in Fig.~\ref{fig:ERT-sbm-vs-fast} the relative ERT difference between the \mlga configurations using fast and standard bit mutation, respectively. As expected, fast mutation performs slightly worse on F1-6, F8, and F11-13. On problems F7, F9, and F15-17, however, fast mutation becomes detrimental to the ERT value for most parameter settings. On problems F10, F14, F18, and F21-25, fast mutation outperforms standard bit mutation, suggesting a potential benefit of pairing the fast mutation with crossover operators to solve more difficult problems. Interestingly, with an increasing $\mu$, the relative ERT of the \mlga quickly shrinks to zero, most notably on F1-6, F8, F9, F11-13.

Interestingly, in~\cite{Buzdalov17fast}, an empirical study has shown that on a randomly generated maximum flow test generation problem, fast mutation is significantly outperformed by standard bit mutation when combined with uniform crossover. Such an observation seems contrary to our findings on F10, F14, F18, and F21-25. However, it is made on a standard $(100+70)$~GA in which both crossover and mutation are applied to the parent in order to generate offspring. We are planning to investigate the effects of this inter-chaining in future work, but this topic is beyond the focus of this study. 

\section{Case-Study: LeadingOnes}
\label{sec:LO}

The surprisingly good performance of the \mlga with $p_c=0.5$ on \leadingones motivates us to investigate this setting in more detail. Before we go into the details of the experimental setup and our results, we recall that for the optimization of \leadingones, the fitness values only depend on the first bits, whereas the tail is randomly distributed and has no influence on the selection. More precisely, a search point $x$ with \leadingones-value $f(x)$ has the following structure: the first $f(x)$ bits are all 1, the $f(x)+1$st bit equals 0, and the entries in the tail (i.e., in positions $[f(x)+2..n]$) did not have any influence on the optimization process so far. For many algorithms, it can be shown that these tail bits are uniformly distributed, see~\cite{Doerr19domi} for an extended discussion.   

\textbf{Experimental setup.} We fix in this section the variator choice to the \textit{offspring-based setting}. We do so because its performance was seen to be slightly better on \leadingones than the population-based choice. We experiment with the parameter settings specified in Sec.~\ref{sec:data}. For each of the settings listed there, we perform  100 independent runs, with a maximal budget of $5n^2$ each.

\begin{figure*}[t]
    \centering
    \setlength{\tabcolsep}{0pt}
    \begin{tabular}{ll}
        \begin{turn}{90}\parbox{.22\textheight}{\centering\scriptsize $(\mu + 1)$\hfill}\end{turn} 
        &\includegraphics[width=.975\linewidth, trim=0mm 28mm 0mm 20mm, clip]{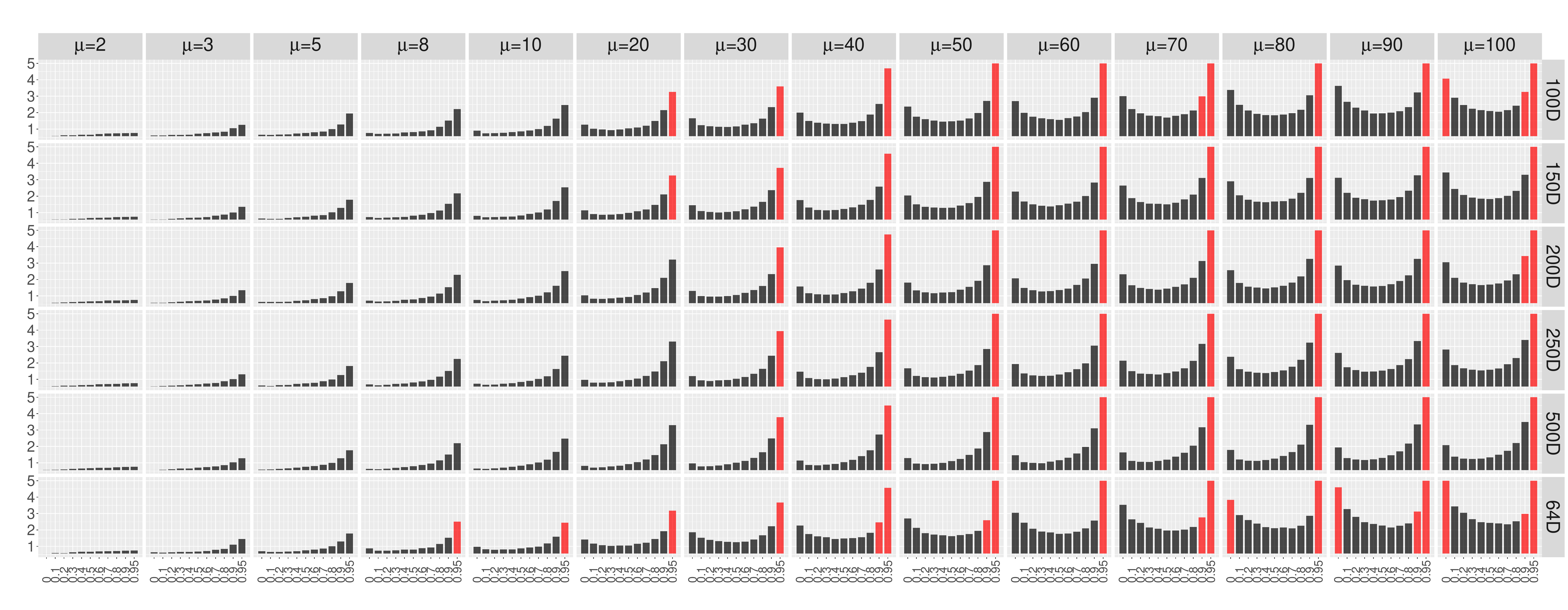} \\
        \begin{turn}{90}\parbox{.22\textheight}{\centering\scriptsize $(\mu + \mu)$\hfill}\end{turn} 
        &\includegraphics[width=.975\linewidth, trim=0mm 2mm 0mm 20mm, clip]{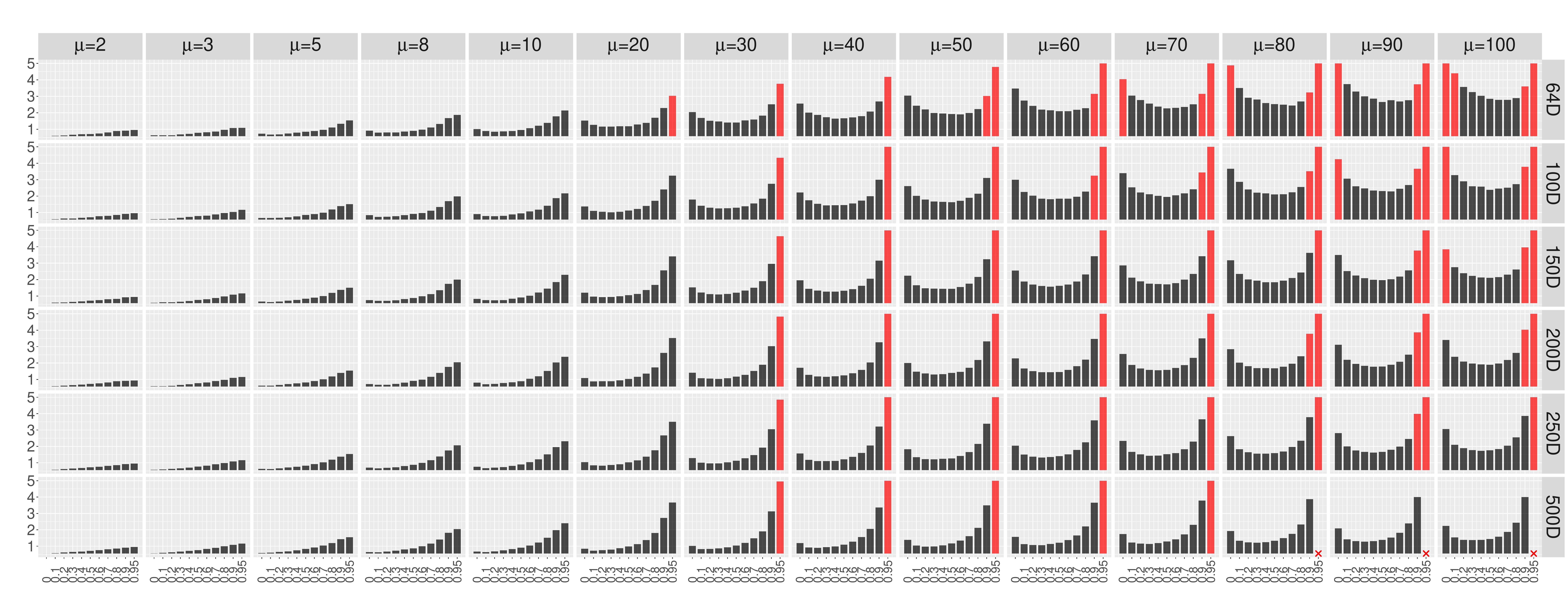}
    \end{tabular}
    \vspace{-10pt}
    \caption{By $n^2$ normalized ERT values for the \mlga using standard bit mutation and uniform crossover on \leadingones, for different values of $\mu$ and for $\lambda =1$ (top) and for $\lambda = \mu$ (bottom). 
    Results are grouped by the value of $\mu$ (main columns), by the crossover probability $p_c$ (minor columns), and by the dimension (rows). The ERTs are computed from 100 independent runs for each setting, with a maximal budget of $5n^2$ fitness evaluations. 
    ERTs for algorithms which successfully find the optimum in all 100 runs are depicted as black bars, whereas ERTs for algorithms with success rates in $(0,1)$ are depicted as red bars. All bars are capped at 5.}
    \label{fig:runtime-mumu-LO-sbm}
\end{figure*}

\textbf{Overall Running Time.} 
We first investigate the impact of the crossover probability on the average running time,  i.e., on the average number of function evaluations that the algorithm performs until it evaluates the optimal solution for the first time. The results for the $(\mu+1)$ and the $(\mu+\mu)$~GA using uniform crossover and standard bit mutation are summarized in Fig.~\ref{fig:runtime-mumu-LO-sbm}. Since not all algorithms managed to find the optimum within the given time budget, we plot as red bars the ERT values for such algorithms with success ratio strictly smaller than $1$, whereas the black bars are reserved for algorithms with $100$ successful runs. All values are normalized by $n^2$, to allow for a better comparison. All patterns described below also apply to the $(\mu + \lceil\mu/2\rceil)$~GA, whose results we do not display for reasons of space. They are also very similar when we replace the mutation operator by the fast mutation scheme suggested in~\cite{doerr2017fast}. 

As a first observation, we note that the pattern of the results are quite regular. As can be expected, the dispersion of the running times is rather small. For reasons of space, we do not describe this dispersion in detail, but to give an impression for the concentration of the running times, we report that the standard deviation of the $(50+1)$~GA on the 100-dimensional \leadingones function is approximately $14\%$ of the average running time across all values of $p_c$.
As can be expected for a genetic algorithm on \leadingones, the average running increases with increasing population size $\mu$, see~\cite{Sudholt13} for a proof of this statement when $p_c=0$.

Next, we compare the sub-plots in each row, i.e., fixing the dimension. We see that the \mlga suffers drastically from large $p_c$ values when $\mu$ is smaller, suggesting that the crossover operator hinders performance. But as $\mu$ gets larger, the average running time at moderate crossover probabilities ($p_c$ around $0.5$) is significantly smaller than that in two extreme cases, $p_c=0$ (mutation-only GAs), and $p_c=0.95$. This observation holds for all dimensions and for both algorithm families, the $(\mu+1)$ and the $(\mu+\mu)$~GA. 

Looking at the sub-plots in each column (i.e., fixing the population size), we identify another trend: for those values of $\mu$ for which an advantage of $p_c>0$ is visible for the smallest tested dimension, $n=64$, the relative advantage of this rate decreases and eventually disappears as the dimension increases.

Finally, we compare the results of the $(\mu+1)$~GA with those of the $(\mu+\mu)$~GA. Following~\cite{JansenJW05}, it is not surprising that for $p_c=0$, the results of the $(\mu+1)$~GA are better than those of the $(\mu+\mu)$~GA (very few exceptions to this rule exist in our data, but in all these cases the differences in average runtime are negligibly small), and following our own theoretical analysis~\cite[Theorem~1]{DoerrYR0B18}, it is not surprising that the differences between these two algorithmic families are rather small: the typical disadvantage of the $(\mu + \lceil\mu/2\rceil)$~GA over the $(\mu+1)$~GA is around 5\% and it is around 10\% for the $(\mu+\mu)$~GA, but these relative values differ between the different configurations and dimensions.  

\textbf{Optimal Crossover Probabilities.} 
To make our observations on the crossover probability clearer, we present in Table~\ref{tab:pc-opt-LO-sbm} a heatmap of the values $p_c^*$ for which we observed the best average running time (with respect to all tested $p_c$ values). We see the same trends here as mentioned above: as $\mu$ increases, the value of $p_c^*$ increases, while, for fixed $\mu$ its value decreases with increasing problem dimension $n$. Here again we omit details for the $(\mu + \lceil\mu/2\rceil)$~GA and for the fast mutation scheme, but the patterns are identical, with very similar absolute values. 

\begin{table}[t]
\centering
\setlength{\tabcolsep}{4pt}
\renewcommand{\arraystretch}{1.2}
\fontsize{7pt}{7pt}\selectfont
\begin{tabular}{|c|r|cccccccccccccc|}
  \cline{2-16}
\multicolumn{1}{c|}{} & \diagbox[width=.8cm, height=0.45cm]{$n$}{$\mu$} & 2 & 3 & 5 & 8 & 10 & 20 & 30 & 40 & 50 & 60 & 70 & 80 & 90 & 100 \\ 
  \cline{1-16}
\multirow[c]{6}{*}{\rotatebox[origin=c]{90}{$(\mu + 1)$}} & 64 & \cellcolor[rgb]{0,0.666666666666667,1}0.0 & \cellcolor[rgb]{0,1,1}0.1 & \cellcolor[rgb]{0,1,1}0.1 & \cellcolor[rgb]{0,1,1}0.1 & \cellcolor[rgb]{0.2,1,0.8}0.2 & \cellcolor[rgb]{0.4,1,0.6}0.3 & \cellcolor[rgb]{0.8,1,0.2}0.5 & \cellcolor[rgb]{0.6,1,0.4}0.4 & \cellcolor[rgb]{0.8,1,0.2}0.5 & \cellcolor[rgb]{0.8,1,0.2}0.5 & \cellcolor[rgb]{1,1,0}0.6 & \cellcolor[rgb]{1,0.666666666666667,0}0.7 & \cellcolor[rgb]{1,1,0}0.6 & \cellcolor[rgb]{1,0.666666666666667,0}0.7 \\ 
& 100 & \cellcolor[rgb]{0,0.666666666666667,1}0.0 & \cellcolor[rgb]{0,1,1}0.1 & \cellcolor[rgb]{0,1,1}0.1 & \cellcolor[rgb]{0,1,1}0.1 & \cellcolor[rgb]{0,1,1}0.1 & \cellcolor[rgb]{0.4,1,0.6}0.3 & \cellcolor[rgb]{0.6,1,0.4}0.4 & \cellcolor[rgb]{0.6,1,0.4}0.4 & \cellcolor[rgb]{0.6,1,0.4}0.4 & \cellcolor[rgb]{0.8,1,0.2}0.5 & \cellcolor[rgb]{0.8,1,0.2}0.5 & \cellcolor[rgb]{0.8,1,0.2}0.5 & \cellcolor[rgb]{0.6,1,0.4}0.4 & \cellcolor[rgb]{1,1,0}0.6 \\ 
& 150 & \cellcolor[rgb]{0,0.666666666666667,1}0.0 & \cellcolor[rgb]{0,1,1}0.1 & \cellcolor[rgb]{0,1,1}0.1 & \cellcolor[rgb]{0,1,1}0.1 & \cellcolor[rgb]{0,1,1}0.1 & \cellcolor[rgb]{0.2,1,0.8}0.2 & \cellcolor[rgb]{0.4,1,0.6}0.3 & \cellcolor[rgb]{0.4,1,0.6}0.3 & \cellcolor[rgb]{0.6,1,0.4}0.4 & \cellcolor[rgb]{0.6,1,0.4}0.4 & \cellcolor[rgb]{0.8,1,0.2}0.5 & \cellcolor[rgb]{0.6,1,0.4}0.4 & \cellcolor[rgb]{0.6,1,0.4}0.4 & \cellcolor[rgb]{0.8,1,0.2}0.5 \\ 
& 200 & \cellcolor[rgb]{0,0.666666666666667,1}0.0 & \cellcolor[rgb]{0,0.666666666666667,1}0.0 & \cellcolor[rgb]{0,1,1}0.1 & \cellcolor[rgb]{0,1,1}0.1 & \cellcolor[rgb]{0,1,1}0.1 & \cellcolor[rgb]{0.2,1,0.8}0.2 & \cellcolor[rgb]{0.2,1,0.8}0.2 & \cellcolor[rgb]{0.4,1,0.6}0.3 & \cellcolor[rgb]{0.4,1,0.6}0.3 & \cellcolor[rgb]{0.4,1,0.6}0.3 & \cellcolor[rgb]{0.6,1,0.4}0.4 & \cellcolor[rgb]{0.6,1,0.4}0.4 & \cellcolor[rgb]{0.6,1,0.4}0.4 & \cellcolor[rgb]{0.6,1,0.4}0.4 \\ 
& 250 & \cellcolor[rgb]{0,0.666666666666667,1}0.0 & \cellcolor[rgb]{0,0.666666666666667,1}0.0 & \cellcolor[rgb]{0,1,1}0.1 & \cellcolor[rgb]{0,1,1}0.1 & \cellcolor[rgb]{0,1,1}0.1 & \cellcolor[rgb]{0.2,1,0.8}0.2 & \cellcolor[rgb]{0.2,1,0.8}0.2 & \cellcolor[rgb]{0.4,1,0.6}0.3 & \cellcolor[rgb]{0.4,1,0.6}0.3 & \cellcolor[rgb]{0.4,1,0.6}0.3 & \cellcolor[rgb]{0.6,1,0.4}0.4 & \cellcolor[rgb]{0.6,1,0.4}0.4 & \cellcolor[rgb]{0.4,1,0.6}0.3 & \cellcolor[rgb]{0.6,1,0.4}0.4 \\ 
& 500 & \cellcolor[rgb]{0,0.666666666666667,1}0.0 & \cellcolor[rgb]{0,0.666666666666667,1}0.0 & \cellcolor[rgb]{0,0.666666666666667,1}0.0 & \cellcolor[rgb]{0,1,1}0.1 & \cellcolor[rgb]{0,1,1}0.1 & \cellcolor[rgb]{0,1,1}0.1 & \cellcolor[rgb]{0,1,1}0.1 & \cellcolor[rgb]{0.2,1,0.8}0.2 & \cellcolor[rgb]{0.2,1,0.8}0.2 & \cellcolor[rgb]{0.4,1,0.6}0.3 & \cellcolor[rgb]{0.4,1,0.6}0.3 & \cellcolor[rgb]{0.4,1,0.6}0.3 & \cellcolor[rgb]{0.4,1,0.6}0.3 & \cellcolor[rgb]{0.4,1,0.6}0.3 \\   
  \cline{1-16}
\multirow[c]{6}{*}{\rotatebox[origin=c]{90}{$(\mu + \mu)$}} & 64 & \cellcolor[rgb]{0,0.666666666666667,1}0.0 & \cellcolor[rgb]{0.2,1,0.8}0.2 & \cellcolor[rgb]{0,1,1}0.1 & \cellcolor[rgb]{0,1,1}0.1 & \cellcolor[rgb]{0.2,1,0.8}0.2 & \cellcolor[rgb]{0.2,1,0.8}0.2 & \cellcolor[rgb]{0.6,1,0.4}0.4 & \cellcolor[rgb]{0.6,1,0.4}0.4 & \cellcolor[rgb]{1,1,0}0.6 & \cellcolor[rgb]{0.8,1,0.2}0.5 & \cellcolor[rgb]{0.8,1,0.2}0.5 & \cellcolor[rgb]{1,0.666666666666667,0}0.7 & \cellcolor[rgb]{0.8,1,0.2}0.5 & \cellcolor[rgb]{1,0.666666666666667,0}0.7 \\ 
& 100 & \cellcolor[rgb]{0,0.666666666666667,1}0.0 & \cellcolor[rgb]{0,0.666666666666667,1}0.0 & \cellcolor[rgb]{0,1,1}0.1 & \cellcolor[rgb]{0,1,1}0.1 & \cellcolor[rgb]{0.2,1,0.8}0.2 & \cellcolor[rgb]{0.4,1,0.6}0.3 & \cellcolor[rgb]{0.4,1,0.6}0.3 & \cellcolor[rgb]{0.4,1,0.6}0.3 & \cellcolor[rgb]{0.8,1,0.2}0.5 & \cellcolor[rgb]{0.6,1,0.4}0.4 & \cellcolor[rgb]{0.8,1,0.2}0.5 & \cellcolor[rgb]{0.8,1,0.2}0.5 & \cellcolor[rgb]{1,1,0}0.6 & \cellcolor[rgb]{0.8,1,0.2}0.5 \\ 
& 150 & \cellcolor[rgb]{0,0.666666666666667,1}0.0 & \cellcolor[rgb]{0,0.666666666666667,1}0.0 & \cellcolor[rgb]{0,1,1}0.1 & \cellcolor[rgb]{0,1,1}0.1 & \cellcolor[rgb]{0.2,1,0.8}0.2 & \cellcolor[rgb]{0.2,1,0.8}0.2 & \cellcolor[rgb]{0.4,1,0.6}0.3 & \cellcolor[rgb]{0.4,1,0.6}0.3 & \cellcolor[rgb]{0.8,1,0.2}0.5 & \cellcolor[rgb]{0.6,1,0.4}0.4 & \cellcolor[rgb]{0.8,1,0.2}0.5 & \cellcolor[rgb]{0.8,1,0.2}0.5 & \cellcolor[rgb]{0.8,1,0.2}0.5 & \cellcolor[rgb]{0.8,1,0.2}0.5 \\ 
& 200 & \cellcolor[rgb]{0,0.666666666666667,1}0.0 & \cellcolor[rgb]{0,0.666666666666667,1}0.0 & \cellcolor[rgb]{0,1,1}0.1 & \cellcolor[rgb]{0,1,1}0.1 & \cellcolor[rgb]{0,1,1}0.1 & \cellcolor[rgb]{0,1,1}0.1 & \cellcolor[rgb]{0.4,1,0.6}0.3 & \cellcolor[rgb]{0.4,1,0.6}0.3 & \cellcolor[rgb]{0.4,1,0.6}0.3 & \cellcolor[rgb]{0.4,1,0.6}0.3 & \cellcolor[rgb]{0.6,1,0.4}0.4 & \cellcolor[rgb]{0.8,1,0.2}0.5 & \cellcolor[rgb]{0.6,1,0.4}0.4 & \cellcolor[rgb]{0.8,1,0.2}0.5 \\ 
& 250 & \cellcolor[rgb]{0,0.666666666666667,1}0.0 & \cellcolor[rgb]{0,0.666666666666667,1}0.0 & \cellcolor[rgb]{0,1,1}0.1 & \cellcolor[rgb]{0,1,1}0.1 & \cellcolor[rgb]{0,1,1}0.1 & \cellcolor[rgb]{0.2,1,0.8}0.2 & \cellcolor[rgb]{0.4,1,0.6}0.3 & \cellcolor[rgb]{0.4,1,0.6}0.3 & \cellcolor[rgb]{0.4,1,0.6}0.3 & \cellcolor[rgb]{0.4,1,0.6}0.3 & \cellcolor[rgb]{0.4,1,0.6}0.3 & \cellcolor[rgb]{0.4,1,0.6}0.3 & \cellcolor[rgb]{0.6,1,0.4}0.4 & \cellcolor[rgb]{0.6,1,0.4}0.4 \\ 
& 500 & \cellcolor[rgb]{0,0.666666666666667,1}0.0 & \cellcolor[rgb]{0,0.666666666666667,1}0.0 & \cellcolor[rgb]{0,0.666666666666667,1}0.0 & \cellcolor[rgb]{0,1,1}0.1 & \cellcolor[rgb]{0,1,1}0.1 & \cellcolor[rgb]{0,1,1}0.1 & \cellcolor[rgb]{0,1,1}0.1 & \cellcolor[rgb]{0.2,1,0.8}0.2 & \cellcolor[rgb]{0.2,1,0.8}0.2 & \cellcolor[rgb]{0.4,1,0.6}0.3 & \cellcolor[rgb]{0.4,1,0.6}0.3 & \cellcolor[rgb]{0.4,1,0.6}0.3 & \cellcolor[rgb]{0.4,1,0.6}0.3 & \cellcolor[rgb]{0.4,1,0.6}0.3 \\ 
   \hline
\end{tabular}
\caption{On \leadingones, the optimal value of $p_c$ for the $(\mu+1)$ and the $(\mu+\mu)$~GA with uniform crossover and standard bit mutation, for various combinations of dimension $n$ (rows) and $\mu$ (columns). Values are approximated from $100$ independent runs each, probing $p_c \in \{0.1k \mid k \in [0..9]\} \cup \{0.95\}$.}
\label{tab:pc-opt-LO-sbm}
\end{table}   

\begin{figure}[!ht]
    \centering
     \includegraphics[width=0.95\linewidth]{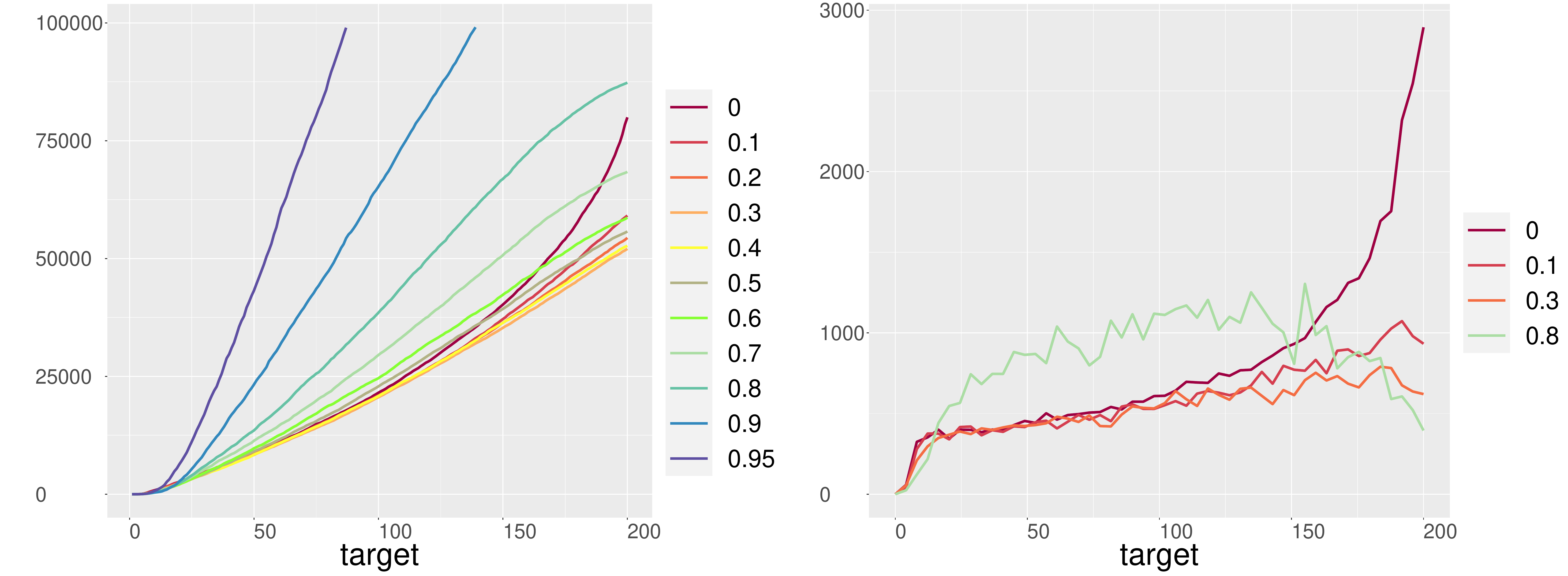}
    \caption{{\small 
    \textbf{Left:} Average fixed-target running times of the $(50+50)$~GA with uniform crossover and standard bit mutation on \leadingones in 200 dimensions, for different crossover probabilities $p_c$. Results are averages of 100 independent runs. \textbf{Right:} Gradient of selected fixed-target curves.}
    }
    \label{fig:LO-FT-200-50-50}
\end{figure}

\textbf{Fixed-Target Running Times.} 
We now study where the advantage of the crossover-based algorithms stems from. We demonstrate this using the example of the $(50+50)$~GA in 200 dimensions. We recall from Table~\ref{tab:pc-opt-LO-sbm} that the optimal crossover probability for this setting is $p_c^*=0.3$. The left plot in Fig.~\ref{fig:LO-FT-200-50-50} is a fixed-target plot, in which we display for each tested crossover probability $p_c$ (different lines) and each fitness value $i\in [0..200]$ ($x$-axis) the average time needed until the respective algorithm evaluates for the first time a search point of fitness at least $i$. The mutation-only configuration ($p_c=0$) performs on par with the best configurations for the first part of the optimization process, but then loses in performance as the optimization progresses. The plot on the right shows the gradients of the fixed-target curves. The gradient can be used to analyze which configuration performs best at a given target value. We observe an interesting behavior here, namely that the gradient of the configuration $p_c=0.8$, which has a very bad fixed-target performance on all targets (left plot), is among the best in the final parts of the optimization. The plot on the right therefore suggests that an adaptive choice of $p_c$ should be investigated further. 

\section{Conclusions}
\label{sec:conclusions}

In this paper, we have analyzed the performance of a family of $(\mu+\lambda)$~GAs, in which offspring are either generated by crossover (with probability $p_c$) or by mutation (probability $1-p_c$). 
On the IOHprofiler problem set, it has been shown that this random choice mechanism reduces the expecting running time on \onemax, \leadingones, and many W-model extensions of those two problems. By varying the value of the crossover probability $p_c$, we discovered on \leadingones that its optimal value $p_c^*$ (with respect to the average running time) increases with the population size $\mu$, whereas for fixed $\mu$ it decreases with increasing dimension $n$.

Our results raise the interesting question of whether a non-asymptotic runtime analysis (i.e., bounds that hold for a fixed dimension rather than in big-Oh notation) could shed new light on our understanding of evolutionary algorithms. We note that a few examples of such analyses can already be found in the literature, e.g., in~\cite{ChicanoSWA15,BuskulicD19}. 
The regular patterns observed in Fig.~\ref{fig:runtime-mumu-LO-sbm} suggest the presence of trends that could be turned into formal knowledge.  

It would certainly also be interesting to extend our study to a \mlga variant using dynamic values for the relevant parameters $\mu$, $\lambda$, crossover probability $p_c$, and mutation rate $p$. We are also planning to extend the study to more conventional \mlga, which apply mutation right after crossover.   

\vspace{1ex}
\textbf{Acknowledgments.}{\footnotesize{ 
Our work was supported by the Chinese scholarship council (CSC No. 201706310143), 
by the Paris Ile-de-France Region, 
and by COST Action CA15140.
}}

}

\end{document}